# Detecting Motifs in System Call Sequences

William O. Wilson, Jan Feyereisl and Uwe Aickelin

School of Computer Science, The University of Nottingham, UK
wow, jqf, uxa@cs.nott.ac.uk

**Abstract.** The search for patterns or motifs in data represents an area of key interest to many researchers. In this paper we present the Motif Tracking Algorithm, a novel immune inspired pattern identification tool that is able to identify unknown motifs which repeat within time series data. The power of the algorithm is derived from its use of a small number of parameters with minimal assumptions. The algorithm searches from a completely neutral perspective that is independent of the data being analysed and the underlying motifs. In this paper the motif tracking algorithm is applied to the search for patterns within sequences of low level system calls between the Linux kernel and the operating system's user space. The MTA is able to compress data found in large system call data sets to a limited number of motifs which summarise that data. The motifs provide a resource from which a profile of executed processes can be built. The potential for these profiles and new implications for security research are highlighted. A higher level system call language for measuring similarity between patterns of such calls is also suggested.

## 1 Introduction

The investigation and analysis of time series data is a popular and well studied area of research. Common goals of time series analysis include the desire to identify known patterns in a time series, to predict future trends given historical information and the ability to classify data into similar clusters. Historically, statistical techniques have been applied to this problem domain whilst Immune System (IS) inspired techniques have remained fairly limited [1]. In this paper we describe the Motif Tracking Algorithm (MTA), a deterministic but non-exhaustive approach to identifying repeating patterns in time series data. The MTA abstracts principles from the human immune system, in particular the immune memory theory of Eric Bell [2]. Implementing principles from immune memory to be used as part of a solution mechanism is of great interest to the immune system community and here we are able to take advantage of such a system. The MTA implements the Bell immune memory theory by proliferating and mutating a population of solution candidates using a derivative of the clonal selection algorithm [3].

A subsequence of a time series that is seen to repeat within that time series is defined as a motif. The objective of the MTA is to find those motifs. The power of the MTA comes from the fact that it has no prior knowledge of the time series to be examined or what motifs exist. It searches in a fast and efficient manner and the flexibility incorporated in its generic approach allows the MTA to be applied across a diverse range of problems. The MTA has already been applied to motif detection in industrial data sets [2]. Here we test its generic properties by applying it to motif identification in system call data sets.

Considerable research has already been performed on identifying known patterns in time series [4]. In contrast little research has been performed on looking for unknown motifs in time series. A distinguishing feature of the MTA is its ability to identify variable length unknown patterns that repeat in a time series. This focus on the detection of unknown patterns makes it an ideal tool for investigating underlying patterns in low level system data generated from the execution of processes on a computer system. The nature of program execution, with the re-use of functions and methods, along with standardised programming structures, implies a set of motifs do exist within each running process. Each process on a system is executed by issuing sequences of system calls which are translated by the kernel into data understandable by the underlying hardware. All processes rely on such system calls making them of high interest to the security community. By looking at the sequences of such system calls, we can observe repeating motifs, which will be identifiable and explainable in terms of higher level functions. The MTA provides an ideal mechanism to compress the data found in these large system call data sets to a limited number of motifs which effectively summarise that data. The motifs could then provide a resource from which we can build a profile of executed processes. These profiles could be used to identify sequences that indicate potentially anomalous process behaviour.

Related work on motif detection is discussed in Section 2 with details of system calls found in Section 3. Terms and definitions used are covered in Section 4 followed by the MTA pseudo code and the problem to be addressed in Sections 5 and 6. Results and future work are found in Sections 7 and 8 before concluding in Section 9.

## 2 Related Work

The search for patterns in data is relevant to a diverse range of fields, including biology, business, finance and computer security. For example, Guan [5] addresses DNA pattern matching using lookup table techniques that exhaustively search the data set to find recurring patterns. Investigations using a piecewise linear segmentation scheme [6] and discrete Fourier transforms [7] provide examples of mechanisms to search a time series for a particular motif of interest. An underlying assumption in all these common approaches is that the pattern to be found is known in advance. The matching task is much simpler as the algorithm just has to find re-occurrences of the required pattern.

The search for unknown motifs is at the heart of the work conducted by Keogh et al. Keoghs probabilistic [8] and Viztree algorithms [9] are very successful in identifying unknown motifs but they require additional parameters compared to the MTA and they also assume prior knowledge of the length of the motif to be found. Motifs longer and potentially shorter than this predefined length may remain undetected in full. Work by Tanaka [10] attempts to address this issue by using minimum description length to discover the optimal length for the motif. Fu et al. [11] use self-organising maps to identify unknown patterns in stock market data, by representing patterns as perceptually important points. This provides an effective solution but again the patterns found are limited to a predetermined length.

This prior awareness of the patterns to find, or their lengths, is not appropriate for intrusion detection systems as by their very nature intrusion techniques are constantly changing to avoid detection. This represents an ideal application for the MTA as it makes no such pre-assumptions and aims to find all unknown motifs of variable length from the data set. Security researchers have been investigating system calls for a number of years. System calls represent a low level communication mechanism between processes and the systems' underlying hardware, providing a high enough level of abstraction for intelligent process behaviour analysis and modelling. The use of system calls for anomaly detection was first introduced by Forrest et al. [12]. Their IS inspired work looks at sequences of system calls using a sliding window mechanism. System calls are used to generate a database of normal behaviour, i.e. self, which is consequently used to capture anomalous behaviour, non-self.

Forrest's work instigated a new stream of intrusion detection research, with some researchers taking the idea of generating a database of normal behaviour and extending it further [13]. Novel approaches to solving related issues by varying the methods and types of signals used to generate a set of normal behaviour were also proposed [14]. Tandon et al. have looked at system call motifs as a source for their normal behaviour profile generation, however this differs from our approach because it uses an exhaustive search mechanism [15].

## 3 Intrusion Detection and System Calls

System calls are lower level functions/methods, which act as a communication channel between higher level processes (e.g. executable commands) and the lower level kernel of an operating system (OS). The system calls perform system actions on behalf of a user. Each system call performs a slightly different atomic task, yet in combination they achieve much more complex functionality. Examples of some of the simplest system calls are the file I/O calls, such as open(), read(), write() and close(). An application or a process can produce on average between a dozen and thousands of system calls per execution, depending on the complexity of the task. As such system calls are ideal data signals from a security point of view, they provide a detailed view of a system or process operation while avoiding complex issues such as encryption or other possible higher level evasion mechanisms.

The focus of the MTA is to look for variable length unknown motifs in the data. This fits nicely with system calls as we are interested in seeing if motifs exist in system call execution sequences. Our inspiration originates from the way that programs are written, compiled and executed. An application usually consists of various classes, objects, methods or functions, along with variables and constants. All these structures are high level constructs, that are processed by lower level libraries which execute appropriate system calls accordingly. These atomic functions, which are deterministic as they do not depend on any variable input, are likely to form the building blocks of our motifs of interest, no matter what application calls them. The combination of such motifs could then provide a resource to generate a profile for a process that distinguishes a permitted execution from a malicious one.

The MTA provides a mechanism for compressing and summarising all this system call data into a number of repeating motifs that are prevalent in the data. The MTA would highlight consistent patterns in the data that are understandable and of value to the user, to aid in the generation of these process profiles. The ability of the MTA to find variable length motifs, with no assumptions about the data or the motifs to find, ensures it is flexible enough to carry out such a task.

## 4 Motif Detection: Terms and Definitions

Whilst immunology provides the inspiration for the theory behind the MTA (see [2] for more information), the work of Keogh et al. [8] is the inspiration for the time series representation used by the MTA. Keogh's Symbolic Aggregate approXimation (SAX) technique for representing a time series was utilised. Many of the following definitions used by the MTA are adapted from the work of Keogh [8], as summarised below.

**Definition 1. Time series.** A time series $T = t_1,...,t_m$ is a time ordered set of m real or integer valued variables. In order to identify patterns in T in a fast and efficient manner we break T up into subsequences.

**Definition 2. Subsequence.** "Given a time series T of length m, a subsequence C of T consists of a sampling of length n ≤ m of contiguous positions from T." [8]. Subsequences are extracted using a sliding window technique.

**Definition 3. Sliding window.** Given a time series T of length m, and a subsequence C of length n, a symbol matrix S of all possible subsequences can be built by sliding a window of size n across T, one point at a time, placing each subsequence into S. After all sliding windows are assessed S will contain (m - n + 1) subsequences. Each subsequence generated could represent a potential match to any of the other subsequences within S. If two subsequences match, we have found a pattern in the time series that is repeated. This pattern is defined as a motif.

**Definition 4. Motif.** A subsequence from T that is seen to repeat at least once throughout T is defined as a motif. The re-occurrence of the subsequence need not be exact for it to be considered as a motif. The relationship between two subsequences $C_1$ and $C_2$ is assessed using a match threshold r. We use the most common distance measure (Euclidean distance) to examine the match between two subsequences $C_1$ and $C_2$, $ED(C_1, C_2)$. If $ED(C_1, C_2) ≤ r$ the subsequences

$C_1$ and $C_2$ are deemed to match and thus are saved as a motif. The motifs prevalent in a time series are detected by the MTA through the evolution of a population of trackers.

Definition 5. Tracker. A tracker represents a signature for a motif sequence that is seen to repeat. It has within it a sequence of 1 to w symbols that are used to represent a dimensionally reduced equivalent of a subsequence. The subsequences generated from the time series are converted into a discrete symbol string using an intuitive technique described in Section 5. The trackers are then used as a tool to identify which of these symbol strings represent a recurring motif. The trackers also include a match count variable to indicate the level of stimulation received during the matching process.

## 5 The Motif Tracking Algorithm

The MTA pseudo code is detailed in Program 1 and a brief summary of this algorithm as applied to system call analysis is described in the subsequent sections. The MTA parameters include the length of a symbol s, the match threshold r, and the alphabet size a.

Convert Time Series T to Symbolic Representation. The MTA takes as input a univariate time series data set consisting of system call data which has been converted to a list of integers as described in Section 6. To minimise amplitude scaling issues with subsequence comparisons across T we normalise the time series. We then use the SAX representation [8] to discretise the time series under consideration. SAX is a powerful compression tool that uses a discrete, finite symbol set to generate a dimensionally reduced version of a time series consisting of symbol strings. This intuitive representation has been shown to rival more sophisticated reduction methods such as Fourier transforms and wavelets [8].

---

Program 1 . MTA Pseudo Code

```
Initiate MTA (s, r, a)
Convert Time series T to symbolic representation
Generate Symbol Matrix S
Initialise Tracker population to size a
While ( Tracker population > 0 )
{
    Generate motif candidate matrix M from S
    Match trackers to motif candidates
    Eliminate unmatched trackers
    Examine T to confirm genuine motif status
    Eliminate unsuccessful trackers
    Store motifs found
    Proliferate matched trackers
    Mutate matched trackers
}
Memory motif streamlining
```
---

Using SAX we slide a window of size s across the time series T one point at a time. Each sliding window represents a subsequence of system calls from T. The MTA calculates the average of the values from the sliding window and uses that average to represent the subsequence.

The MTA now converts this average value into a symbol string. The user predefines the size a of the alphabet used to represent the time series T. Given T has been normalised we can identify the breakpoints for the alphabet characters that generate a equal sized areas under the Gaussian curve [8]. The average value calculated for the sliding window is then examined against the breakpoints and converted into the appropriate symbol. This process is repeated for all sliding windows across T to generate m-s+1 subsequences, each consisting of symbol strings comprising one character.

**Generate Symbol Matrix S**. The string of symbols representing a subsequence is defined as a word. Each word generated from the sliding window is entered into the symbol matrix S. The MTA examines the time series T using these words and not the original data points to speed up the search process. Symbol string comparisons can be performed efficiently to filter out bad motif candidates, ensuring the computationally expensive Euclidean distance calculation is only performed on those motif candidates that are potentially genuine.

Having generated the symbol matrix S, the novelty of the MTA comes from the way in which each generation a selection of words from S, corresponding to the length of the motif under consideration, are extracted in an intuitive manner as a reduced set and presented to the tracker population for matching.

**Initialise Tracker Population to Size a**. The trackers are the primary tool used to identify motif candidates in the time series. A tracker comprises a sequence of 1 to w symbols. The symbol string contained within the tracker represents a sequence of symbols that are seen to repeat throughout T. Tracker initialisation and evolution is tightly regulated to avoid proliferation of ineffective motif candidates. The initial tracker population is constructed of size a to contain one of each of the viable alphabet symbols predefined by the user. Each tracker is unique, to avoid unnecessary duplication of solution candidates and wasted search time.

Trackers are created of a length of one symbol. The trackers are matched to motif candidates via the words presented from the stage matrix S. Trackers that match a word are stimulated; trackers that attain a stimulation level ≥ 2 indicate repeated words from T and become candidates for proliferation. Given a motif and a tracker that matches part of that motif, proliferation enables the tracker to extend its length by one symbol each generation until its length matches that of the motif.

**Generate Motif Candidate Matrix M from S**. The symbol matrix S contains a time ordered list of all the words, each containing just one symbol, that are present in the time series. Neighbouring words in S contain significant overlap as they were extracted via the sliding windows. Presenting all words in S to the tracker population would result in potentially inappropriate motifs being identified between neighbouring words. To prevent this issue such 'trivial' match candidates are removed from the symbol matrix S. Trivial match elimination (TME) is achieved as a word is only transferred from S for presentation to the tracker population if it differs from the previous word extracted. This allows the MTA to focus on significant variations in the time series and prevents excessive time being wasted on the search across uninteresting variations.

Excessively aggressive trivial match elimination is prevented by limiting the maximum number of consecutive trivial match eliminations to s, the number of data points encompassed by a symbol. In this way a subsequence can eliminate as trivial all subsequences generated from sliding windows that start in locations contained within that subsequence (if they generate the same symbol string) but no others. The reduced set of words selected from S is transferred to the motif candidate matrix M and presented to the tracker population for matching.

**Match Trackers to Motif Candidates.** During an iteration each tracker is taken in turn and compared to the set of words in M. Matching is performed using a simple string comparison between the tracker and the word. We define a match to occur if the comparison function returns a value of 0, indicating a perfect match between the symbol strings. Each matching tracker is stimulated by incrementing its match counter by 1.

**Eliminate Unmatched Trackers.** Trackers that have a match count $>1$ indicate symbols that are seen to repeat throughout T and are viable motif candidates. Eliminating all trackers with a match count $< 2$ ensures the MTA only searches for motifs from amongst these viable candidates. Knowledge of possible motif candidates from T is therefore carried forward by the tracker population. After elimination the match count of the surviving trackers is reset to 0.

**Examine T to Confirm Genuine Motif Status.** The surviving tracker population indicates which words in M represent viable motif candidates. However motif candidates with identical words may not represent a true match when looking at the time series data underlying the subsequences comprising those words. In order to confirm whether two matching words X and Y, containing the same symbol strings, correspond to a genuine motif we need to apply a distance measure to the original time series data associated with those candidates. The MTA uses the Euclidean distance to measure the relationship between two motif candidates ED(X,Y).

If ED(X,Y) $\leq$ r a motif has been found. The match count of that tracker is stimulated to indicate a match. A memory motif is created to store the symbol string associated with X and Y. The start locations of X and Y are also saved. For further information on the derivation of this matching threshold please refer to [2]. The MTA then continues its search for motifs, focusing only on those words in M that match the surviving tracker population in an attempt to find all occurrences of the potential motifs. The trackers therefore act as a pruning mechanism, reducing the potential search space to ensure the MTA only focuses on viable candidates.

**Eliminate Unsuccessful Trackers.** The MTA now removes any unstimulated trackers from the tracker population. These trackers represent symbol strings that were seen to repeat but upon further investigation with the underlying data were not proven to be valid motifs in T.

**Store Motifs Found.** The motifs identified during the confirmation stage are stored in the memory pool for review. Comparisons are made to remove any duplication. The final memory pool represents the compressed representation of the time series, containing all the re-occurring patterns present.

**Proliferate Matched Trackers.** Proliferation and mutation are needed to extend the length of the tracker so it can capture more of the complete motif. At the end of the first generation the surviving trackers, each consisting of a word containing a single symbol, represent all the symbols that are applicable to the motifs in T. The complete motifs in T can only consist of combination of these symbols. This subset of trackers is therefore stored as the mutation template for use by the MTA.

Proliferation and mutation to lengthen the trackers will only involve symbols from the mutation template and not the full symbol alphabet, as any other mutations would lead to unsuccessful motif candidates. During proliferation the MTA takes each surviving tracker in turn and generates a number of clones equal to the size of the mutation template. The clones adopt the same symbol string as their parent.

**Mutate Matched Trackers.** The clones generated from each parent are taken in turn and extended by adding a symbol taken consecutively from the mutation template. This creates a tracker population with maximal coverage of all potential motif solutions and no duplication. The tracker pool is fed back into the MTA ready for the next generation. A new motif candidate matrix M consisting of words with two symbols is now formulated to present to the evolved tracker population. In this way the MTA builds up the representation of a motif one symbol at a time each generation to eventually map to the full motif.

Given the symbol length s we generate a word consisting of two consecutive symbols by taking the symbol from matrix S at position i and that from position i+s. Repeating this across S, and applying trivial match elimination as per Section 5, the MTA obtains a new motif candidate matrix M in generation two, each entry of which contains a word of two symbols, covering a length of 2 x s.

The MTA continues to prepare and present new motif candidate matrix data to the evolving tracker population each generation. The motif candidates are built up one symbol at a time and matched to the lengthening trackers. This flexible approach enables the MTA to identify unknown motifs of a variable length. This process continues until all trackers are eliminated as non matching and the tracker population is empty. Any further extension to the tracker population will not improve their fit to any of the underlying motifs in T.

Memory Motif Streamlining. The MTA streamlines the memory pool, removing duplicates and those encapsulated within other motifs to produce a list of motifs that it associates with T.

## 6 Detection of System Call Patterns

This paper demonstrates that an execution of a process shall produce a sequence of system calls containing a number of motifs of variable lengths and these shall be identifiable by the MTA. Such motifs should re-occur when the same or similar processes are run. The motifs in system call sequences can be used in various security applications, for example as a data reduction tool for behaviour profiling within an Intrusion Detection System (IDS).

In our experiments we have two machines connected by a local network. The client machine (Windows XP machine running an SSH client PuTTY version 0.57) connects to the server machine (Debian Linux) which then performs actions based on commands sent by the client. Our experiments use a VMware virtual machine, running a Debian Linux distribution, v.3.3.5-13, with a Linux kernel, v.2.4.27-2, as our SSH server. The SSH daemon process (OpenSSH 3.8.1p1 Debian-8.sarge.6) is monitored along with all its children, using the standard strace utility. All system calls generated by the SSH daemon and its child processes are logged and stored in separate files based on their process ID (PID).

The following sequence of actions is executed to generate our data set. The client connects to the server and an SSH session is established. The following commands, chosen at random, are then issued by the client: ls, ls, ls -lsa, pwd, ls, ps, ps aux, ls -lsa, chmod a+x file, chmod a+x directory, ls -lsa, chmod a-x file. The client then disconnects from the SSH server.

The PID files generated are concatenated to produce one file. During concatenation the PID file from the child with the smallest PID is added to the parent data, this is then repeated for each remaining child process. Concatenation results in a single data set containing system call names with their respective arguments. The data is further pre-processed by converting the individual system call names to their appropriate Linux OS id numbers and removing their arguments. This generates a one dimensional data set comprising a sequence of 8,040 system call numbers.

From the data set generated it can be observed that a small sequence of system calls relating to one particular monitored child process is repeated across a large proportion of the data. This process looks after the SSH terminal operation for the duration of the whole SSH session. Repeated read() and write() calls and various real time system call actions are performed over and over again. Due to the basic nature of this repeating sequence, it is not deemed of interest to our analysis. Instead we focus on the last 1,000 system calls from the data set, to in-vestigate motifs that occur during the last seven commands issued by the client. The data set generated is available at http://cs.nott.ac.uk/~ jqf/MTA_scdata.dat.

## 7 Results

Having introduced the MTA we now provide some experimental results which examine the ability of the MTA to identify motifs present in system call data. As defined in Section 6 the data set examined consists of 1,000 system calls represented by an ordered list of 1,000 integers. A bind threshold $r = 0$ was set since system call sequences need to match identically. Symbol length s and alphabet size a values were varied to investigate the sensitivity of the MTA to these parameters. The MTA was written in C++ and run on a Windows XP machine with a Pentium M 1.7 Ghz processor with 1.0 Gb of RAM.

### 7.1 System Call Motifs Identified by the MTA

In this scenario $a = 10$ to give a large alphabet diversity and s took the values 10, 20 and 40. To evaluate the impact on speed and accuracy the MTA was run with trivial match elimination (TME) and with no trivial match elimination (NTME). We focus on significant motifs whose length exceeds 40 system calls to enable fair comparison across different values of s. With $s = 10$ and NTME eight motifs are identified. Table 1 lists these motifs with the number of system calls they encompass and the start locations where they occur in the data set.

Motif 1 dominates the data set, it consists of 280 system calls and occurs twice in the data set from locations 386 and 717. Figure 1 presents the list of system calls from location 350 to 1,000 and motif 1 is clearly evident in this sequence. From the commands issued during the SSH session (Section 6), we observe the existence of motifs within the command list itself as there are repetitions of the ls and chmod commands. From the MTA's analysis of the system call data set, motif 1 relates to the repetition of these two observed commands. The ls commands contain the same arguments (-lsa) across both repetitions, whilst the chmod command includes execute permissions for all users to a file in the first occurrence and removes those same privileges from the file during the second occurrence. Motif 1 represents two processes occurring in succession at two different time points within the overall SSH session. Motif 1 also contains other sub-motifs such as motif 7. Motif 7 occurs at three different positions as seen from Table 1. This motif represents the chmod command that was executed three times during the session, each time with different arguments. The motifs found by the MTA are a super-set of those evident in the original command list, validating the accuracy of the MTA. The existence of motif 7 shows that applications with varying arguments (i.e. performing different actions), have atomic motifs that could be used for data reduction in a security application.

Table 1. List of motifs found by the MTA

| Motif No. | Length | Start Locations |
|---|---|---|
| 1 | 280 | 386, 717 |
| 2 | 80 | 0, 227 |
| 3 | 70 | 8, 160, 235 |
| 4 | 50 | 198, 262 |
| 5 | 50 | 668, 950 |
| 6 | 40 | 39, 191, 266, 324 |
| 7 | 40 | 619, 668, 950 |
| 8 | 40 | 77, 120 |

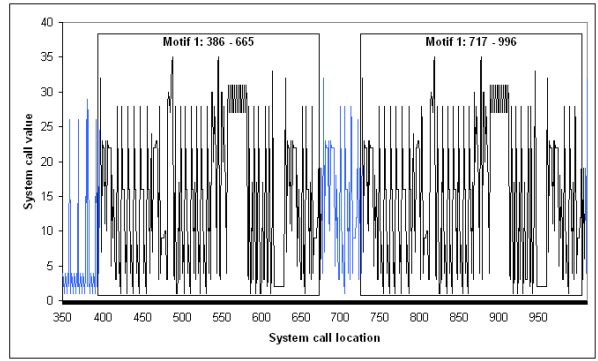

Fig. 1. Illustration of system calls 350 to 1,000, highlighting the occurrence of Motif 1 from system call 386 to 665 and 717 to 996

Motif 2 is the second largest motif, which has two repetitions in the data set. It relates to the execution of the ps command. This command occurred twice in succession, with different arguments, spawning a new process each time. Motif 2 represents 80 system calls that are identical across these two spawned processes. Motifs 3, 4 and 8 partly overlap with motif 2 indicating a subset of system calls from motif 2 that is consistent across all these motifs. However this subset occurs with a higher frequency than motif 2, representing similarities between atomic parts of the ps processes not fully captured by the motif 2. This highlights parts of the process execution that are more dynamic and input dependent and which need to be dealt with when considering an IDS.

Motif 6 again relates to a component of the ps command. At a lower level, the ps command reads a number of small files from the /proc/ directory of the Linux OS and prints the read information onto the screen. This information shows the running processes on the system to the user. From this a recurring sub-pattern of open(), read() and close() system calls with various arguments is observed. This sub-pattern is largely dependant on the input of the ps command. In this case it is the number of running processes on the monitored system, which in turn is the number of files in the aforementioned /proc/ directory. The motifs generated by the MTA are formed by such atomic system call sequences. The randomness in the input, which results in motifs of variable lengths is of major importance when considering IDS systems. For this purpose a system call expression language is proposed which, besides giving the MTA an alternative representation to assess system call similarity, gives security researchers a regular expression type language for describing system call motifs at a higher level. This language is described in more detail in Section 8.

In the above analysis we have focused on significant motifs with sequences exceeding 40 system calls. The advantage of using the MTA on system call data and not the command list becomes apparent when we look at the shorter motifs that are generated, as these indicate atomic motifs that are found across varying command instructions. One such motif, referred to as Z, has a length of 30 system calls and occurred five times at positions 387, 620, 669, 718 and 951. Motif Z relates to the re-occurrence of a sequence that occurs during the initialisation of newly spawned processes executed across the commands ls -lsa, chmod a+x file, chmod a+x directory, ls -lsa and chmod a-x file respectively. Each C application under Linux, when it starts, calls and loads the standard C library, libc. The occurrences of motif Z corresponds to the loading of this library. Thus the MTA has found a motif that is present but embedded across differing commands. This approach of using system calls as input to the MTA, and not the commands, would aid in the detection of exploits that are far smaller than the commands themselves. An example of such an exploit is the SQL slammer worm which is only 376 bytes long, compared to the text segment of the chmod command of 29,212 bytes.

### 7.2 Sensitivity to Changes in the Symbol Length s

In total 961 of the 1,000 system calls are identified as being part of one or more of these eight motifs. From these results it is clear the MTA is able to successfully identify a reduced set of motifs from the large system call data set. By varying the value of s and the use of trivial match elimination we can examine the sensitivity of the MTA and assess its ability to retain knowledge of these eight motifs. The results of this sensitivity analysis can be seen in Table 2.

Table 2 shows the total number of motifs found and the execution time of the MTA for the various values of s. In addition a measure of the quality of the motifs found is included by multiplying the identified length of the motif by the identified frequency and summing for all motifs found. Any omissions in the length or frequency of the complete motif will cause a decline in this quality measure.

Table 2. Sensitivity of the MTA to variations in the symbol length s

| s | Motifs found | | Execution time (sec) | | Motif quality measure | |
|---|---|---|---|---|---|---|
| | NTME | TME | NTME | TME | NTME | TME |
| 10 | 8 | 6 | 315.8 | 262.0 | 1,490 | 1,230 |
| 20 | 5 | 4 | 56.9 | 26.5 | 1,140 | 940 |
| 40 | 4 | 3 | 12.1 | 1.9 | 960 | 720 |

As the symbol length s increases, the number of motifs detected declines. This appears logical as a higher s implies the search is less fine grained. Introducing TME also reduces the number of motifs found. TME significantly reduces the size of the motif candidate matrix M resulting in fewer candidates being examined. TME is key to the dimensionality reduction of the original data set leading to a fast search process, however it would appear that its inclusion does lead to a loss in detection accuracy.

Given NTME, the MTA is only able to identify four of the eight motifs (1, 2, 6 and 7) if s rises from 10 to 40. The quality measure also indicates that, of the motifs found, there appears to be a loss in the detection of the full length or frequency of occurrence. The quality measure falls from 1,490 to 960. Of the four motifs still detected we lose 40 system calls from motif 1 and we only detect two of the three repetitions of motif 7. However raising the symbol length from 10 to 40 results in a 96.2% reduction in the MTA execution time, taking only 12.1 seconds compared to 315.8.

## 7.3 Sensitivity to Changes in the Alphabet Size a

Adjusting the alphabet size alters the symbol set used to represent the time series. Reducing a means a greater diversity of sequences are now grouped together as similar. TME with a reduced alphabet set should lead to a larger number of trivial match eliminations, leading to a faster but potentially less accurate search. This hypothesis is confirmed when we look at Table 3 which lists the motifs found for various alphabet sizes. In this scenario $s = 20$, $r = 0$ and a took the values 10, 8, 6, and 4.

Table 3 shows the alphabet size has no impact on the detection ability of the MTA if there is NTME. The five motifs detected when $s = 20$, $a = 10$ (Table 2) are always found and have the same quality measurement. However the search time of the MTA with NTME improves by 30.1% as a is reduced from 10 to 4.

With TME activated, changes to a have a more significant impact on the motifs detected. Reducing a from 10 to 8 causes the MTA to lose track of motif 8 but it now finds motif 2. As motif 2 is longer than motif 8 we get an improvement in the overall motif quality measure from 940 to 1,140. Reducing a further from 8 to 4 causes the MTA to lose motif 6 but gain knowledge of motif 4. Thus we see that TME causes a change in the location in search space where the MTA conducts its search, resulting in less consistent results.

Table 3. Sensitivity of the MTA to variations in the alphabet size a

| a | Motifs found | | Execution time (sec) | | Motif quality measure | |
|---|---|---|---|---|---|---|
| | NTME | TME | NTME | TME | NTME | TME |
| 10 | 5 | 4 | 56.9 | 26.5 | 1,140 | 940 |
| 8 | 5 | 5 | 48.0 | 19.8 | 1,140 | 1,140 |
| 6 | 5 | 3 | 43.3 | 16.1 | 1,140 | 860 |
| 4 | 5 | 5 | 39.8 | 6.3 | 1,140 | 1,140 |

One could imply that this inconsistency due to trivial match elimination is detrimental to this particular search problem but this need not be the case. As is evident from Table 3 trivial match elimination significantly improves the search time of the MTA and the results from including trivial match elimination are still satisfactory. When a=4 activating trivial match elimination results in the MTA still finding five motifs but it reduces the search time by 84.2% from 39.8 seconds to 6.3 seconds and with no loss to the quality of those five motifs found.

### 7.4 Summary Discussion of Results

From these results it is apparent that the MTA is able to identify motifs that are present in this system call data set. The MTA can compress the original data set of 1,000 system calls down to eight repeating motifs. A trade off between speed and accuracy becomes apparent as the user is able to adjust the parameters of the algorithm to speed up the search process at the cost of a reduction in detection capability, allowing a flexible search mechanism.

The sensitivity to changes in s and a noted here is due to the nature of system call data. In this paper we group system calls together as similar by averaging their system call values over a fixed sized window. These are then grouped by boundary conditions and represented by a symbol which is then subject to trivial match elimination. With system calls there is no real relationship between two separate system call values, i.e. system call 2 is not twice as large as system call 1. Therefore one could argue that a more appropriate representation may be more suitable as is discussed in Section 8.

The results show that the MTA, developed to identify motifs in financial and industrial data sets, is successful in identifying motifs in system call data due to its generic and flexible approach. It provides a useful tool to compress a large data set into small subset of repeating patterns that are of immediate value to the user.

## 8 Future Work

The difficulty with analysing system call sequences for the purposes of intrusion detection is that the variety of sequences generated is largely dependant on the diversity of the application's input. This potential variety sidesteps most forms of pattern detection as long as the detection mechanism is not able to encode the variations in a manner that is granular enough to be able to distinguish between normal and anomalous patterns. To address this issue, as part of our future work we propose a system call expression language (SCEL), which acts as a higher level regular expression type language consisting of constructs representing atomic system call motifs of meaningful actions.

An example of this language can be presented using motif Z of length 30 as described in Section 7. In the SCEL motif Z could be represented by a higher level construct, such as lib_loading(libc). Where lib_loading represents a particular set of motifs for that action and (libc) denotes the class to which those motifs belong. Similar constructs could be devised for other operations which contain atomic motifs, representative of a higher level functionality. The motifs 2, 3, 4, and 8 in Table 1 indicate the re-occurrence of the open(), read() and close() system calls. These three calls could now be represented as one file_read(small) construct to be used for files below a certain size threshold. In addition a file_read(large) construct containing a wild card for the number of read() system calls between the open() and close() calls can be generated for instances when reading larger files, where numerous read() calls are executed depending on the file's size. When reading such a file an attribute of the language construct could denote the number of motifs present in an observed process.

For example lib_loading(libc)[1], file_read(small)[*], other(*)[*] could denote a complete ps command being executed. This language tool would prove of value to a user as it focuses on a high level of abstraction while maintaining the ability to conduct fine grained analysis of system calls. This new representation for system call similarity could now also be used as input for the MTA to enhance its motif detection ability.

## 9 Conclusion

The search for patterns or motifs in data represents a generic problem area that is of great interest to a huge variety of researchers. By extracting motifs that exist in data we gain some understanding as to the nature and characteristics of that data, so that we can benefit from that knowledge. The motifs provide an obvious mechanism to cluster, classify and summarise the data, placing great value on these patterns.

Little research has been performed looking for unknown motifs in time series. The MTA takes up this challenge using a novel immune inspired approach to evolve a population of trackers that seek out and match motifs present in a time series. The MTA uses a minimal number of parameters with minimal assumptions and requires no knowledge of the data examined or the underlying motifs, unlike other alternative approaches.

In this paper the MTA was applied to motif detection in system call data. The MTA was shown to compress the data set into a limited number of motifs that provide good coverage of the original data set resulting in a minimal loss of information. The authors propose that these motifs highlight repeating or atomic functions that can be used to build profiles of "system behaviour". These profiles could then assist in tasks such as anomaly detection or behaviour classification.

The authors provide information on a system call expression language that addresses system call granularity issues for computer security applications in the future. In its current form we believe the MTA offers a valuable contribution to an area of research that at present has received surprisingly little attention.